\documentclass[times,twocolumn,final]{elsarticle}

\usepackage{framed,multirow}

\usepackage{amssymb}
\usepackage{latexsym}

\usepackage{url}
\usepackage[table]{xcolor}
\definecolor{newcolor}{rgb}{.8,.349,.1}

\usepackage{amsmath, amsfonts, bbm}
\usepackage{algorithm, algorithmic}
\usepackage{tikz}
\usepackage{hyperref}
\usepackage{booktabs}
\usepackage{multirow}
\usepackage{hhline}
\usepackage{todonotes}
\usetikzlibrary{bayesnet}
\usepackage{amsthm}
\newtheorem{theorem}{Theorem}[section]

\newtheorem{lemma}[theorem]{Lemma}
\DeclareMathOperator*{\argmax}{arg\,max}

\usepackage{subcaption}
\usepackage{multirow}
\usepackage{graphicx}

\journal{Pattern Recognition Letters}

\begin{document}

\clearpage

\ifpreprint
  \setcounter{page}{1}
\else
  \setcounter{page}{1}
\fi

\begin{frontmatter}

\title{Topic-aware latent models for representation learning on networks}

\author[1]{Abdulkadir \c{C}elikkanat}\corref{cor1}
\ead{abdcelikkanat@gmail.com}
\author[1]{Fragkiskos D. Malliaros}\corref{cor2}
\ead{fragkiskos.malliaros@centralesupelec.fr}

\address[1]{Paris-Saclay University, CentraleSupélec, Inria, Centre for Visual Computing, 91192 Gif-Sur-Yvette, France}

\begin{abstract}
Network representation learning (NRL) methods  have received significant attention over the last years thanks to their success in several graph analysis problems, including node classification, link prediction and clustering. Such methods aim to map each vertex of the network into a low dimensional space in a way that the structural information of the network is preserved.  Of particular interest are methods based on random walks; such methods  transform the  network into a collection of node sequences, aiming  to learn  node representations by predicting the context of each node within the sequence. In this paper, we introduce TNE, a generic framework to enhance the embeddings of nodes acquired by means of random walk-based approaches with topic-based information. Similar to the concept of topical word embeddings in Natural Language Processing, the proposed model first assigns each node to a latent community  with the favor of various statistical graph models and community detection methods, and then learns the enhanced topic-aware representations. We evaluate our methodology in two downstream tasks: node classification and link prediction. The experimental results demonstrate that by incorporating node and community embeddings, we are able to outperform widely-known baseline NRL models.
\end{abstract}


\end{frontmatter}


\section{Introduction}\label{intro}
Graphs are important mathematical structures commonly used to represent  objects and their relations in real-world systems such as the World Wide Web, social networks, and biological networks. Of particular importance is how to deal with learning tasks on graphs, such as the ones of friendship recommendation in social networks and protein function prediction in protein-protein interaction networks---with the major challenge being how to incorporate information about the structure of the graph in the learning process.
To this direction, a plethora of approaches have emerged under the area of \textit{network representation learning} (NRL) \citep{DBLP:journals/debu/HamiltonYL17}. The main goal of NRL models is to learn feature vectors corresponding to the nodes of the graph (also known as \textit{node embeddings}), by preserving important structural properties of the network; those vectors can later be used to perform various analysis and mining tasks including visualization, node classification and link prediction with the favor of machine learning algorithms.

Initial studies in the field of network representation learning, have mostly relied on \textit{matrix factorization} techniques, since various properties and  interactions between nodes can be expressed as matrix operations. Nevertheless, such methods are not scalable to large-scale networks mainly due to their increased running time complexity---especially for graphs consisting of millions of nodes and edges \citep{DBLP:journals/debu/HamiltonYL17}. More recent studies have concentrated on developing methods suitable for relatively large-scale networks --  being able to effectively approximate the underlying objective functions that capture meaningful information about the nodes of the graph and their properties.

Many node representation learning methods have been inspired by the advancements in the area of \textit{natural language processing} (NLP), borrowing various ideas originally developed for computing \textit{word embeddings}. A prominent example here is the \textit{Skip-Gram} architecture \citep{word2vec}, which aims to find latent representations of words by estimating their context within the sentences of a textual corpus. To this direction, many pioneer studies in NRL \citep{deepwalk, node2vec} utilize the idea of random walks to transform graphs into a collection of sentences -- as an analogy to the area of natural language -- and these sentences or walks are later being used to learn node embeddings. 
 
Although random walk-based approaches are strong enough to capture  local connectivity patterns, they mainly suffer to sufficiently convey information about more global structural properties. More precisely, real-world networks have an inherent clustering (or community) structure, which can be utilized to further improve the predictive capabilities of node embeddings. One can interpret such structural information based on an analogy to the concept of \textit{topics} in a collection of documents. In a similar way as word embeddings can be enhanced with topic-based information \citep{topical_word}, here we aim at empowering node embeddings by employing information about the latent community structure of the network---that can be achieved by a process similar to the one of \textit{topic modeling}.

In this paper, we propose \textit{Topical Node Embeddings} (TNE), a framework in which node embeddings are enhanced with topic (or community) information towards learning topic-aware node representations---something that leads to further improvements in the performance on downstream tasks. Local clustering patterns are of great importance on many applications, enabling to better grasp hidden information of the network. For instance, consider two individuals sharing common friends or interests, that are not yet represented by a direct link in the network. A careful analysis of the local community structure can further help to infer the missing information (e.g., missing links), and therefore boost the predictive capabilities of node embeddings. Motivated by that, the proposed TNE framework aims to directly leverage the latent community structure of the graph while learning node embedding vectors. The main contributions of the paper can be summarized as follows:

\begin{itemize}
    \item \textit{Latent graph models and topic representations}. We show how existing latent space discovery models, such as community detection and topic models, can be incorporated in the node representation learning process.
    
	\item \textit{Node representation learning framework}. We propose a new model, called \textsc{TNE}, which first learns community embeddings from the graph, and then uses them to improve the node representations extracted by random walk-based methods. We examine various instances of this model, studying their properties.
	
	\item \textit{Enriched feature vectors}. We perform a detailed empirical evaluation of the embeddings learned by \textsc{TNE} on the tasks of node classification and link prediction. As the experimental results indicate, the proposed model learn feature vectors which can boost the performance on downstream tasks.

\end{itemize}

The rest of the paper is organized as follows. Section \ref{sec:related_work} describes the related work, and in Section \ref{sec:background}, we give the fundamental concepts for unfamiliar readers and formulate the problem. In Section \ref{sec:topic_repr_learning}, we describe the concept of topical node representation learning. The proposed TNE model is presented in Section \ref{sec:topical_node_embedding}. Section \ref{sec: experiments} presents the experimental results, and finally, in Section \ref{sec: conclusions} we conclude our work providing also future research directions. 


\section{Related Work}\label{sec:related_work} 
Several methods have been proposed to learn latent node representations  in an unsupervised manner \citep{DBLP:journals/debu/HamiltonYL17, 8294302}.  Traditional unsupervised feature learning methods typically aim at factorizing a matrix representation, chosen in a way to properly consider  the underlying properties of a given network. Characteristic examples here constitute models that  preserve  first-order proximity of nodes, such as  Laplacian Eigenmaps \citep{laplacian_eigenmap} and \textsc{IsoMap} \citep{isomap}. More recently,   algorithms including LINE \citep{line},  \textsc{GraRep} \citep{grarep} and \textsc{HOPE} \citep{hope} were designed to preserve higher-order proximities of nodes. Nevertheless, despite the fact that matrix factorization approaches offer an elegant way to capture the desired properties, they mainly suffer from their time complexity. 

Random walk-based methods \citep{DBLP:journals/debu/HamiltonYL17}, including \textsc{Node2Vec} \citep{node2vec}, \textsc{DeepWalk} \citep{deepwalk}, have gained considerable attention, mainly due their elegant way to define proximity among nodes based on random walks, as well the computational efficiency of the \textit{Skip-Gram} model. Following this line of research, various extensions have been proposed \citep{random_walk_ddrw,random_walk_struc2vec,harp,biasedwalk,DBLP:conf/www/EpastoP19,efge-aaai20,kernelne}. Lastly, it was recently shown that that \textsc{DeepWalk} and \textsc{Node2Vec}  implicitly perform matrix factorizations \citep{implicit_factorization,netsmf}. 

\begin{figure}[t]
\centering
\includegraphics[scale=0.5]{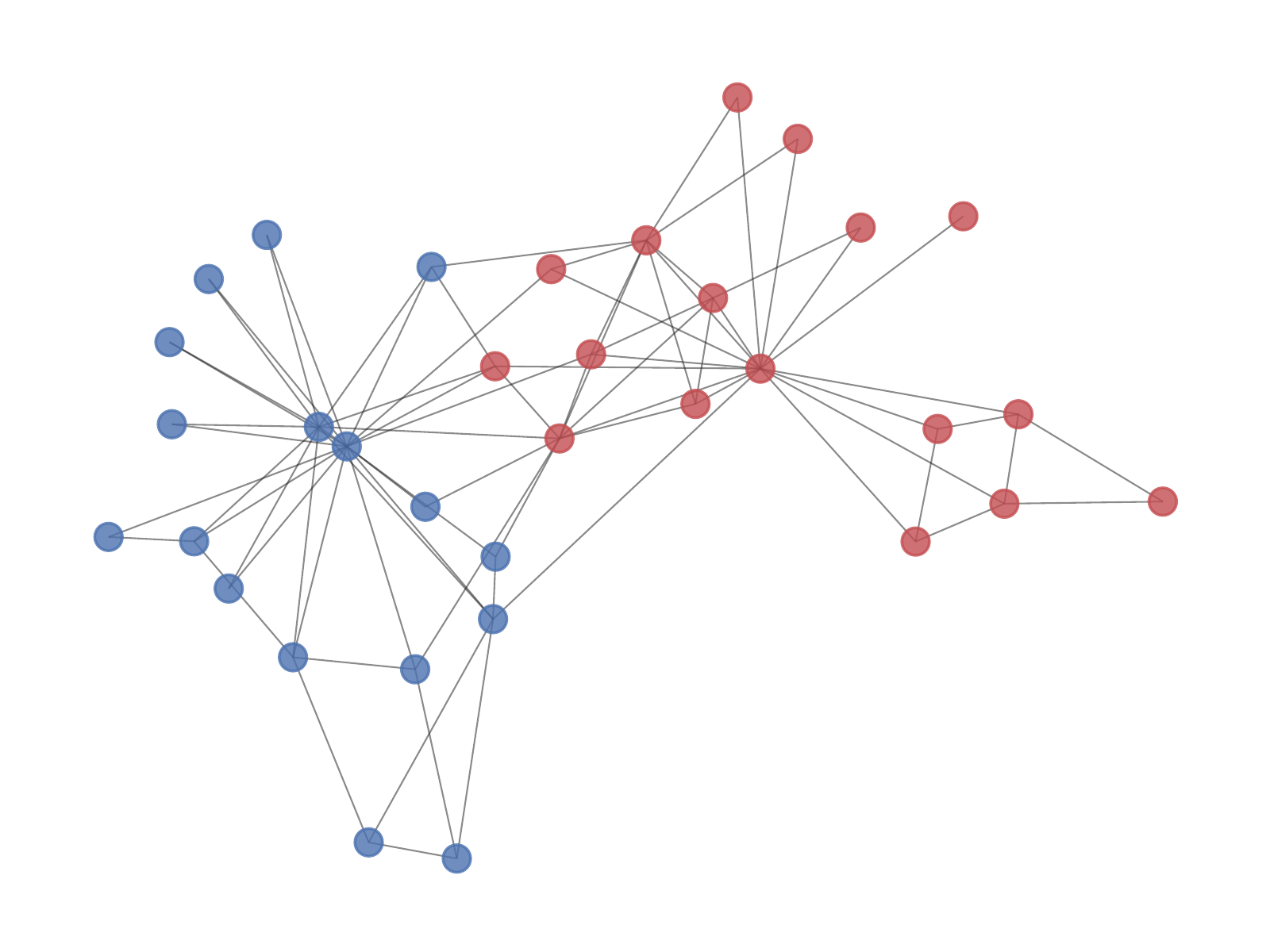}
\caption{The topic-community assignments in Zachary's karate club network. Each node $v$ is assigned to a community label $z$ maximizing the posterior probability $Pr(z|v)$ by \textsc{TNE-Glda} model. \label{fig:karate}}
\end{figure}

To the best of our knowledge, very  few models  benefit from the community structure of real networks while learning embeddings. 
The \textsc{ComE} model \citep{come} proposes a closed-loop procedure among the encoding of communities, learning node embeddings and community detection in the network. \textsc{M-NMF} \citep{DBLP:conf/aaai/WangCWP0Y17} targets to learn node representations by incorporating community structure information in a non-negative matrix factorization formulation. \textsc{COSINE} \citep{Zhang2018COSINECS} is a generative model learning the social network embeddings from information diffusion cascades. A recent approach, \textsc{GEMSEC} \citep{gemsec}, learns node embeddings with an explicitly defined community preserving objective function.
As we will present shortly, our work aims at independently learning  node and community (topic) embeddings, and then combining them into expressive topical feature vectors.

\section{Background Concepts}\label{sec:background}
We will use $G=(\mathcal{V}, \mathcal{E})$ to denote a graph, where $\mathcal{V}$ is the set of nodes and $\mathcal{E}$ indicates the set of edges.  Our goal is to find a mapping function $\beta:\mathcal{V} \rightarrow \mathbb{R}^{\mathcal{D}}$, where $\beta[v]$ will correspond to the representation of node $v$ in a lower-dimensional space $\mathbb{R}^{\mathcal{D}}$; $\mathcal{D}$ is generally referred to as the embedding or dimension size, and is much smaller than the cardinality of the vertex set, $|\mathcal{V}|$.

Node embedding methods based on the popular \textit{Skip-Gram}  architecture (e.g., \citep{node2vec, deepwalk, biasedwalk}) learn node representations using node sequences produced by random walks over a given network. They mainly target to maximize the likelihood of the occurrences of nodes within a certain distance with respect to each other, as follows:

\begin{equation}\label{eq: node_emb_obj_func}
    \!\!\!\!\mathcal{F}_N(\alpha, \beta) \!:= \!\argmax_{\Omega}\!\!\!\sum_{\textbf{w}\in\mathcal{W}}\sum_{1\leq i \leq L} \sum_{-\gamma \leq j\not=0 \leq \gamma }\!\!\!\!\! \log Pr(v_{i+j} | \ v_i; \ \Omega),\!
\end{equation}

\noindent where $\mathcal{W}$ is the set of random walks $\textbf{w}=(v_1,\ldots,v_{i},\ldots,v_{L})\in\mathcal{V}^{L}$, $\Omega=(\alpha, \beta)$ is the model parameters that we would like to learn and $\gamma$ refers to the window size. Any nodes appearing inside the window size of \textit{center} node $v_i$ are referred to as \textit{context} nodes. Note that, we obtain two different representation vectors $\alpha[v]$ and $\beta[v]$ for each node $v \in \mathcal{V}$, one per role of the node as center and context; in the experimental evaluation, we will only consider vector $\beta[v]$, which corresponds to the vector if the node is interpreted as a center node.

\section{Learning Topic Representations}\label{sec:topic_repr_learning}

Complex networks, such as those arising from social or biological settings, consist of latent clusters of different sizes in which the nodes are more likely to be connected to each other \citep{newman_community_structure, Li_2020, 8936884}. Our main goal here is to use the latent clusters of a network in order to obtain enriched representations. This can be achieved by enhancing node embedding vectors with \textit{topic  representations}. By replacing a node $v_i$ with its community label $z_i$ in a random walk, we learn \textit{community embeddings} by predicting the nodes in the context of a community label. More formally, we can define our objective function to learn topic representations as follows:

\begin{equation}\label{eq: community_emb_obj_func}
\!\!\!\! \mathcal{F}_T(\tilde{\alpha}, \tilde{\beta})\! :=\! \argmax_{\widetilde{\Omega}} \!\!\!\sum_{\textbf{w} \in \mathcal{W}}\!\sum_{1 \leq i \leq L} \sum_{-\gamma \leq j\not=0 \leq \gamma}\!\!\!\!\! \log Pr\left(v_{i+j} \ | \ z_i; \ \widetilde{\Omega} \right).\!
\end{equation}

\noindent By maximizing the log-probability above, we obtain the embedding vectors corresponding to each community label $z_i \in \{1,\ldots,\mathcal{K}\}$, where $\mathcal{K}$ indicates the number of latent communities. In this work, we mainly use two approaches to detect latent communities. The first one is based on novel combination of generative statistical models accompanied with random walks, while the second one is based on traditional community detection algorithms that utilize the network structure itself. 

\subsection{Random walks and generative graph models} \label{sec:topic}
Most real-world networks can be expressed as a combination of nested or overlapping communities \citep{overlap_networks}. Therefore, when a random walk is initialized, it does not only visit neighboring nodes but also traverses communities in the network (see Fig. \ref{fig:complete}b). In that regard, we assume that each random walk can be represented as random mixtures over latent communities, and each community can be characterized by a distribution over nodes. In other words, we can write the following generative model for each walk over the network:

\begin{enumerate}
	\item For each $k \in \{1,\ldots,\mathcal{K}\}$
	\begin{itemize}
		\item $\phi_k \sim Dir(b_0)$
		\end{itemize}
	\item For each walk $\textbf{w}=(v_1,\ldots,v_i,\ldots,v_L)$
	\begin{itemize}
		\item $\theta_{\textbf{w}} \sim Dir(a_0)$
		\item For each vertex $v_i \in \boldsymbol{w}$
		\begin{itemize}
			\item $z_{i} \sim Multinomial(\theta_{\textbf{w}})$
			\item $v_i \sim Multinomial(\phi_{z_{i}})$
		\end{itemize}
	\end{itemize}
\end{enumerate}

\noindent Here, $\mathcal{N}$ is the number of walks and $\mathcal{L}$ is the walk length.

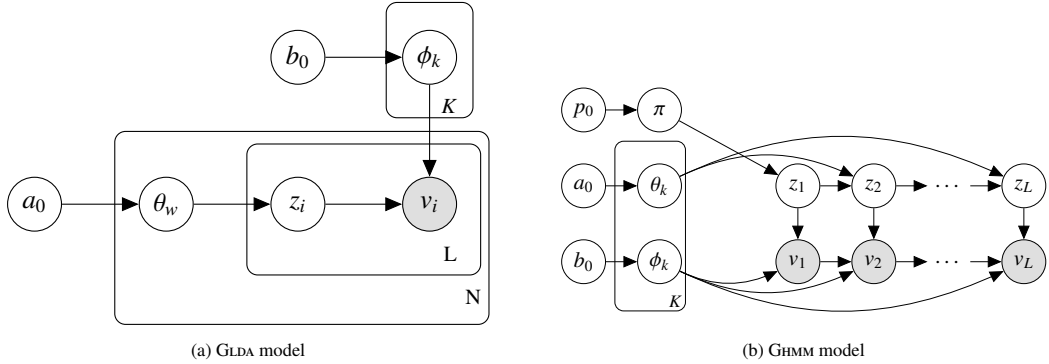
\begin{figure*}
\centering
  \begin{subfigure}[b]{0.4\linewidth}
  \centering
    \resizebox{1.0\textwidth}{!}{
\begin{tikzpicture}
\node[obs] (v) {$v_i$} ; %
\node[latent, left=of v] (z) {$z_i$} ; %
\node[latent, left=of z] (theta) {$\theta_w$} ; %
\node[latent, left=of theta] (alpha) {$a_0$} ; %
\node[latent, above=1.2cm of v] (phi) {$\phi_{k}$} ; %
\node[latent, left=of phi] (beta) {$b_0$} ; %

\plate[inner sep=0.3cm,  yshift=0.12cm] {plate1} {(z) (v)} {L}; %
\plate[inner sep=0.2cm,  yshift=-0.05cm, xshift=-3.0] {plate2} {(theta) (plate1)} {N}; %
\plate[inner sep=0.2cm, yshift=0.15cm, ] {plate3} {(phi)} {$K$}; %
\edge {alpha} {theta} ; %
\edge {theta} {z} ; %
\edge {beta} {phi} ; %
\edge {z} {v} ; %
\edge {phi} {v} ; 
\end{tikzpicture}
}
    \caption{\textsc{Glda} model}
    \label{glda}
  \end{subfigure}
  \quad\quad
  \begin{subfigure}[b]{0.4\linewidth}
  \centering
	\resizebox{1.0\textwidth}{!}{
	\begin{tikzpicture}
		\node[latent] (beta) {$a_0$} ;
		\node[latent, right=0.5cm of beta] (phi) {$\theta_{k}$} ;
		\node[latent, below=0.5cm of beta] (alpha) {$b_0$} ; 
		\node[latent, below=0.5cm of phi] (A) {$\phi_k$} ; 
		\node[latent, right=1.5cm of phi] (z1) {$z_1$} ; 
		\node[obs, below=0.5cm of z1] (v1) {$v_1$} ;
		\node[latent, right=0.5cm of z1] (z2) {$z_2$} ; 
		\node[obs, below=0.5cm of z2] (v2) {$v_2$} ;
		\node[right=0.5cm of z2] (zdots) {$\dots$} ; 
		\node[right=0.5cm of v2] (vdots) {$\dots$} ;
		\node[latent, right=0.5cm of zdots] (zl) {$z_L$} ; 
		\node[obs, right=0.5cm of vdots] (vl) {$v_L$} ;
		
		\node[latent, above=0.5cm of beta] (p) {$p_0$} ;
		\node[latent, right=0.5cm of p] (pi) {$\pi$} ;
		
		\plate[inner sep=0.2cm, yshift=0.15cm, xshift=-0.15cm] {plate1} {(phi) (A)} {$K$}; 
		
		\path[every node/.style={font=\sffamily\small}]
		(p) edge[->] node [left] {} (pi)

		(beta) edge[->] node [left] {} (phi)
		(alpha) edge[->] node [left] {} (A)
		
		(z1) edge[->] node [left] {} (z2)
		(z2) edge[->] node [left] {} (zdots)
		(zdots) edge[->] node [left] {} (zl)
		
		(v1) edge[->] node [left] {} (v2)
		(v2) edge[->] node [left] {} (vdots)
		(vdots) edge[->] node [left] {} (vl)
		
		(z1) edge[->] node [left] {} (v1)
		(z2) edge[->] node [left] {} (v2)
		(zl) edge[->] node [left] {} (vl)
		
		(pi) edge[->] node [left] {} (z1)
		(phi) edge[->, bend left=25] node [left] {} (z2)
		(phi) edge[->, bend left=25] node [left] {} (zl)
		(A) edge[->, bend right=25] node [left] {} (v1)
		(A) edge[->, bend right=25] node [left] {} (v2)
		(A) edge[->, bend right=25] node [left] {} (vl);
	\end{tikzpicture}
	}
  \caption{\textsc{Ghmm} model}\label{fig:ghmm}
  \end{subfigure}
  \caption{Plate representations of random walk-based topic representation models.}
\end{figure*}

If we consider each random walk as a document and the collection of random walks as a corpus, it can be seen that the statistical process defined above corresponds to the well known Latent Dirichlet Allocation (LDA) model \citep{lda}. Therefore, each community corresponds to a distinct topic in the terminology of NLP (we use the terms \textit{topic} and \textit{community} interchangeably in the rest of the paper). We will refer to this model as \textsc{Glda} (the plate representation is shown in Fig. \ref{glda}).  As we show in Lemma \ref{lemma: occurences}, the relative frequency of the occurrences of a node in the generated walks is proportional to its degree in the network for large number of walks or walk lengths.
This property was first empirically demonstrated in the work of \cite{deepwalk}, allowing \textit{Skip-Gram} models to be applied on real-world graphs. Here we provide a formal argument of this empirical observation.

\begin{lemma}\label{lemma: occurences}
Let $G=(\mathcal{V}, \mathcal{E})$ be a connected graph, and $\{X_l\}_{l\geq 1}$ be a Markov chain with  state space $\mathcal{V}=\{1,\ldots,N\}$ and  transition matrix $P$, where $P_{ij}$ is defined as $1 \slash d_i$ for each edge $(i, j) \in \mathcal{E}$ and $0$ otherwise. If the Markov chain is aperiodic, then

\begin{equation*}
    \begin{aligned}
    \lim_{L \to \infty}\frac{1}{L}\mathbb{E}\left[O_L^i\right] = \frac{d_i}{2|\mathcal{E}|},
    \end{aligned}
\end{equation*}

\noindent where $O^{i}_L$ is a random variable representing the number of occurrences of the node $i$ in a random walk of length $L$.
\end{lemma}
\begin{proof}
Since the graph is connected, each state can be accessed by any other one. Thus, the Markov chain is also irreducible having a unique limiting distribution $\boldsymbol{\pi}$. Note that $\pi_i$ is equal to $d_i / \sum_{k \in \mathcal{V}}d_k$ since it satisfies $\boldsymbol{\pi}P = \boldsymbol{\pi}$. Then, we can write:
\begin{equation*}
\begin{aligned}
    \lim_{L \rightarrow \infty} \frac{1}{L} \mathbb{E}\left[O^i_L \right] \!&=\!  \lim_{L \rightarrow \infty} \frac{1}{L} \mathbb{E}\left[ \sum_{l=1}^L\mathbbm{1}_{\{X_l=i\}} \right] \!=  \!\!\lim_{L \rightarrow \infty} \frac{1}{L} \sum_{l=1}^L\!\mathbb{E}\left[ \mathbbm{1}_{\{X_l=i\}} \right] \\
    \!&=\! \lim_{L \rightarrow \infty} \frac{1}{L} \sum_{l=1}^L Pr\left(X_l=i\right) = \pi_i = \frac{d_i}{\sum_{k \in \mathcal{V}}d_k}\\
    \!&=\! \frac{d_i}{2|\mathcal{E}|},
\end{aligned}
\end{equation*}
where the equality in the second line follows from \textit{Ces\`aro theorem} \citep{devito2007harmonic}, since $Pr(X_l=i)$ converges to $\pi_i$ as $l$ goes to infinity.
\end{proof}

In the previous \textsc{Glda} model, the latent community assignment of each node is independently chosen from the community label of the previous node in the random walk. However, the hidden state of the current node can play an important role towards determining the next vertex to visit, as the random walk also traverses through communities. Therefore, we can write the following generative process, by modifying the \textsc{Glda} model:

\begin{enumerate}
	\item For each $k \in \{1,\ldots,\mathcal{K}\}$
	\begin{itemize}
		\item $\phi_k \sim Dir(b_0)$
		\item $\theta_{k} \sim Dir(a_0)$
	\end{itemize}
	\item $\pi \sim Dir(p_0)$
	\item For each walk $\textbf{w}=(v_1,\ldots,v_i,\ldots,v_L)$
	\begin{itemize}
	    \item $z_{1} \sim Dir(\pi)$
		\item For each vertex $v_i \in \boldsymbol{w}$, for all $i<L$
		\begin{itemize}
		    \item $v_{i} \sim Multinomial(\phi_{z_{i}})$
			\item $z_{i+1} \sim Multinomial(\theta_{z_{i}})$
		\end{itemize}
		\item $v_{L} \sim Multinomial(\phi_{z_{L}})$
	\end{itemize}
\end{enumerate}

\noindent The above model, in fact, corresponds to the well-known \textit{Hidden Markov Model} (HMM) with symmetric Dirichlet priors over transition and emission distributions. In our experiments, we adopt the \textit{Infinite Hidden Markov Model} (\textsc{IHMM}) \citep{beam_sampling} (we will refer to this model as \textsc{Ghmm}; plate representation is shown in Fig. \ref{fig:ghmm}). Note that, unlike the \textsc{Glda} model, in the generation of each node sequence the same transition probabilities are used. Also,  vectors $\theta_k$ and $\phi_k$ contain $\mathcal{K}$ and $|\mathcal{V}|$ components, respectively. 

\subsection{Network structure-based modeling} \label{sec:community}
In the previous models, the generated random walks are used to detect the community (or topic) assignment of each node in the given node sequence. Here, we utilize two additional community detection models, which directly target to extract communities of nodes from a given network.  The first one corresponds to the well-known \textsc{Louvain} algorithm by \cite{louvain} that extracts communities based on modularity maximization, while the second one to the \textsc{BigClam} model for overlapping community detection \citep{bigclam}. 

\section{Topical Node Embeddings}\label{sec:topical_node_embedding}

\begin{figure*}
    \centering
	\includegraphics[width=.99\textwidth]{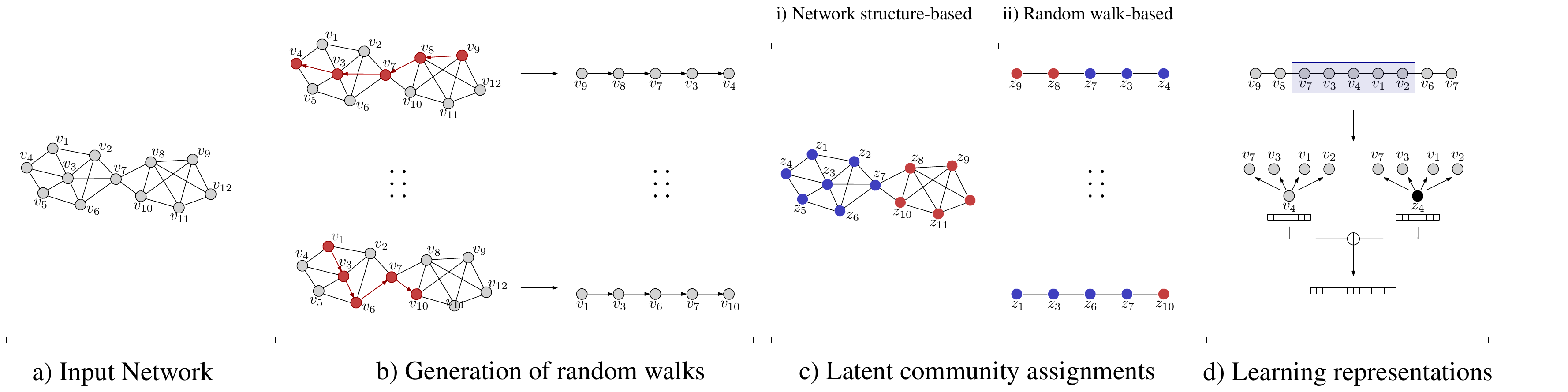}
	\caption{Schematic representation of the \textsc{TNE} model. The final representations are learned by combining node and topic embeddings. A node's representation is learned using random walks performed over the network; its topic representation is similarly learned by assigning a topic/community label based either on random walks (\textsc{TNE-Glda}, \textsc{TNE-Ghmm}) or network structure-based approaches (\textsc{TNE-Louvain}, \textsc{TNE-BigClam}).} \label{fig:complete}
\end{figure*}

In this section, we will describe the proposed Topical Node Embeddings (TNE) model for learning topic-aware node representations.
An overview of the model is given in Fig. \ref{fig:complete}. TNE aims to enhance node embeddings using information about the underlying topics of the graph obtained by the models described in Section \ref{sec:topic_repr_learning}.  This can be achieved by learning node and topic embedding vectors independently of each other, jointly maximizing the objectives defined in Eq. \eqref{eq: node_emb_obj_func} and \eqref{eq: community_emb_obj_func}. Combining these two objectives, we derive the following:

\begin{equation*}\label{eq:tne_obj_func}
\begin{aligned}
\max_{\Omega, \widetilde{\Omega}} \sum_{\textbf{w} \in \mathcal{W}}\sum_{v_i \in \textbf{w}}\sum_{-\gamma \leq j\not=0 \leq \gamma}\left[ \log Pr\left(v_{i+j} | v_i; \Omega \right) + \log Pr\left(v_{i+j} | z_i; \widetilde{\Omega}\right) \right].
\end{aligned}
\end{equation*}

\noindent \textit{Skip-Gram} models the probability measure in the above equation using the \textit{softmax} function:
\begin{align*}
Pr\left(v_{i+j}|v_i\right) := \frac{\exp\left(\alpha[v_{i+j}]^{\top} \cdot \beta[v_i]\right)}{\sum_{u\in V}\exp\left(\alpha[u]^{\top} \cdot \beta[v_i]\right)}.
\end{align*}

\noindent In our approach, though, we use the \textit{sigmoid} function by adopting the negative sampling strategy  \citep{word2vec}, in order to make our computations more efficient. After obtaining the node and topic representations, our final step is to efficiently incorporate these two feature vectors, $\beta[v]$ and $\tilde{\beta}[z]$ of node $v$ and community label $z$ respectively, so as to obtain the final topic-enhanced node embedding. For this purpose, we concatenate node embedding vector $\beta[v]$ with the expected topic vector with respect to the distribution $p(\cdot|v_i)$. Our strategy can be formulated  as follows: 

\begin{align}
    \beta[v_i] \oplus \sum_{k\in\{1,\ldots,\mathcal{K}\}}Pr(k|v)\cdot\tilde{\beta}[k],
    \label{eq:conc}
\end{align}

\noindent where $\oplus$ indicates the concatenation operation. We refer to the final vector obtained after concatenating the node and topic feature vectors as \textit{topical node embedding}. Algorithm \ref{alg:our_model} provides the pseudocode of the proposed TNE model. 
\renewcommand{\algorithmicrequire}{\textbf{Input:}}
\renewcommand{\algorithmicensure}{\textbf{Output:}}

\begin{algorithm}[h]
\begin{algorithmic}[1]
\REQUIRE Graph $G=(\mathcal{V},\mathcal{E})$; number of walks: $\mathcal{N}$; walk length: $\mathcal{L}$; window size: $\gamma$; number of communities: $\mathcal{K}$, topic representation learning method: $T$; node embedding size: $\mathcal{D}_n$;  community embedding size: $\mathcal{D}_t$ 
\ENSURE Embedding vectors of length $\mathcal{D}_n + \mathcal{D}_t$
\STATE Perform $\mathcal{N}$ random walks of length $\mathcal{L}$ for each node.
\STATE Learn node representations by optimizing Eq. \eqref{eq: node_emb_obj_func}.
\STATE Learn topic representations by optimizing Eq. \eqref{eq: community_emb_obj_func}, using any of the models $T$ of Sec. \ref{sec:topic_repr_learning}.
\STATE Concatenate node and topic embeddings with Eq. \eqref{eq:conc}.
\end{algorithmic}
\caption{Topical Node Embeddings (TNE)}
\label{alg:our_model}
\end{algorithm}

The general structure of our framework follows. First, we need a collection of walks over the network to learn node and topic embeddings; here we have utilized biased random walks, similar to \textsc{Node2Vec}. We further produce node-context pairs and use \textit{Skip-Gram} to learn node embeddings following Eq. \eqref{eq: node_emb_obj_func}.  Then, we choose a strategy (shown as $T$ in Alg. \ref{alg:our_model}) to learn topic representations. This step is quite flexible in the formulation of TNE.  One approach is to generate topic assignments $z_i$ of each node $v_i \in \mathcal{V}$ in the walk $w \in \mathcal{W}$, based on the random walk-based generative graph models \textsc{Glda} and \textsc{Ghmm},  defined in Sec \ref{sec:topic}. Alternatively, we can directly infer the latent clustering structure based on \textsc{BigClam} and \textsc{Louvain} models, as described in Sec.  \ref{sec:community}.  Lastly, we combine  node and topic embeddings using Eq. \eqref{eq:conc} to obtain the final topical node embedding vectors. Depending on the method used to learn topical representations, we will refer to the corresponding instances of TNE as \textsc{TNE-Glda}, \textsc{TNE-Ghmm}, \textsc{TNE-Louvain} and \textsc{TNE-BigClam}.

\subsection{Running time complexity analysis}
The time complexity of \textsc{TNE} varies depending on the algorithm used to detect latent topics and communities. The node and topic representations can be learned in $\mathcal{O}\left(\mathcal{D}\cdot|V|\cdot\mathcal{S}\right)$ time \citep{verse} using the \textit{negative sampling} technique  for a given node sequences and topic assignments, where $\mathcal{S}$ is the number of samples. If the \textsc{Louvain} algorithm is chosen, the communities can be detected in $\mathcal{O}(|V|\cdot\log^2|V|)$ operations, while $\mathcal{O}\left(|V|\cdot\mathcal{K}\right)$ steps are required for \textsc{BigClam} when the number of communities is high. The models that rely on random walks and generative models (\textsc{Glda}, \textsc{Ghmm}), can be run in $\mathcal{O}\left(\mathcal{N}\cdot L \cdot \mathcal{K} \cdot I\right)$, where $I$ is the number of iterations \citep{fast_collapsed_lda, beam_sampling}.

\section{Experimental Evaluation} \label{sec: experiments}
In this section, we firstly present the experimental set-up used in our study, describing the baseline methods and datasets used in the evaluation. We also provide details about parameter settings for \textsc{TNE} and baselines. The performance of the proposed model is examined in two downstream tasks: node classification and link prediction. Our model has been implemented in Python and the source code can be found at: \url{https://abdcelikkanat.github.io/projects/TNE/}.

\subsection{Baseline methods}
We evaluate the performance of \textsc{TNE} against seven widely used baseline methods. $(i)$ \textsc{DeepWalk} applies \textit{Skip-Gram} \citep{deepwalk} with uniform random walks. $(ii)$ \textsc{Node2Vec} \citep{node2vec} extends \textsc{DeepWalk} following biased random walks.  $(iii)$ \textsc{LINE} \citep{line} aims at learning representations relying on first-order and second-order node proximity. $(iv)$ \textsc{HOPE} generates embedding vectors by capturing higher order information of the network; in our experiments we have used the \textit{Katz} index. $(v)$ \textsc{NetMF} \citep{implicit_factorization} is a matrix factorization approach based on the pointwise mutual information of node co-occurrences. $(vi)$ \textsc{GEMSEC} \citep{gemsec} is a random walk model that learns the community structure and the embeddings simultaneously. $(vii)$ Finally, \textsc{M-NMF} \citep{DBLP:conf/aaai/WangCWP0Y17} learns embeddings via a modularity-preserving matrix factorization scheme.
	
\subsection{Datasets}\label{subsec: datasets}
We have used seven different networks in our experiments. \textsl{CiteSeer} \citep{harp} and \textsl{Cora} \citep{cora} are  citation networks  constructed using articles as nodes, with edges representing citations. \textsl{DBLP} is a co-authorship graph, where an edge exists between nodes if two authors have co-authored at least one paper. The labels used for classification represent the research areas. \textsl{AstroPh} and \textsl{HepTh} are both collaboration networks built from the papers submitted to the  \textit{ArXiv} repository for Physics related topics. \textsl{Facebook} \citep{facebook} is a social network obtained from a survey conducted via a Facebook application. Lastly, \textsl{Gnutella} \citep{gnutella} is the peer-to-peer file sharing network. In all cases, we consider networks as undirected to ensure the consistency of the experiments. Table \ref{tbl:data-stats} provides more detailed information about the datasets.

\begin{table}[]
\caption{Statistics of networks. $|\mathcal{V}|$: number of nodes, $|\mathcal{E}|$: number of edges, $|\Lambda|$: number of labels and $|\mathcal{C}|$: number of connected components.}
\label{tbl:data-stats}
\resizebox{\linewidth}{!}{%
\begin{tabular}{rcccccl}
\toprule
\multicolumn{1}{l}{} & \textbf{$|\mathcal{V}|$} & \textbf{$|\mathcal{E}|$} & \textbf{$|\Lambda|$} & \textbf{$|\mathcal{C}|$} & \textbf{Avg. Degree} &  \textbf{Type} \\\midrule
\textsl{CiteSeer} & 3,312 & 4,660 & 6 & 438 & 2.814 & Citation \\
\textsl{Cora} & 2,708 & 5,278 & 7 & 78 & 3.898  & Citation \\
\textsl{DBLP} & 27,199 & 66,832 & 4 & 2,115 & 4.914  & Co-authorship\\\midrule
\textsl{AstroPh} & 17,903 & 19,7031 & - & 1 & 22.010 &  Collaboration \\
\textsl{HepTh} & 8,638 & 24,827 & - & 1 & 5.7483 & Collaboration\\
\textsl{Facebook}  & 4,039 & 88,234 & - & 1 & 43.6910 &  Social\\
\textsl{Gnutella} & 8,104 & 26,008 & - & 1 & 6.4186& Peer-to-peer \\\bottomrule
\end{tabular}%
}
\end{table}

\subsection{Parameter settings}\label{subsec: paramsetting}
In this section, we  describe the parameters' settings that we have used for our experiments and clarify the strategies that we follow. 
To be consistent with the related literature, we have considered  walks of length $\mathcal{L}=10$, number of walks $\mathcal{N}=80$ and window size $\gamma=10$ for \textsc{GEMSEC} and for all random walk-based approaches. The remaining parameters of the baseline methods are set to their default values with the embedding size of $128$. The difference instances of \textsc{TNE} are fed with biased random walks similar to those used in  \textsc{Node2Vec}, setting hyper-parameters $p$, $q$ to $1.0$.  To speed up the training process, we adopt the negative sampling \citep{word2vec} strategy. Stochastic Gradient Descent has been used for optimization, setting the initial learning rate to $0.0025$ and its minimum value to $10^{-5}$. We learn topic and node embedding vectors of sizes $32$ and $96$, respectively, so as the obtain feature vectors of length $128$.

\begin{figure}[H]
\centering
    \includegraphics[width=0.9\linewidth]{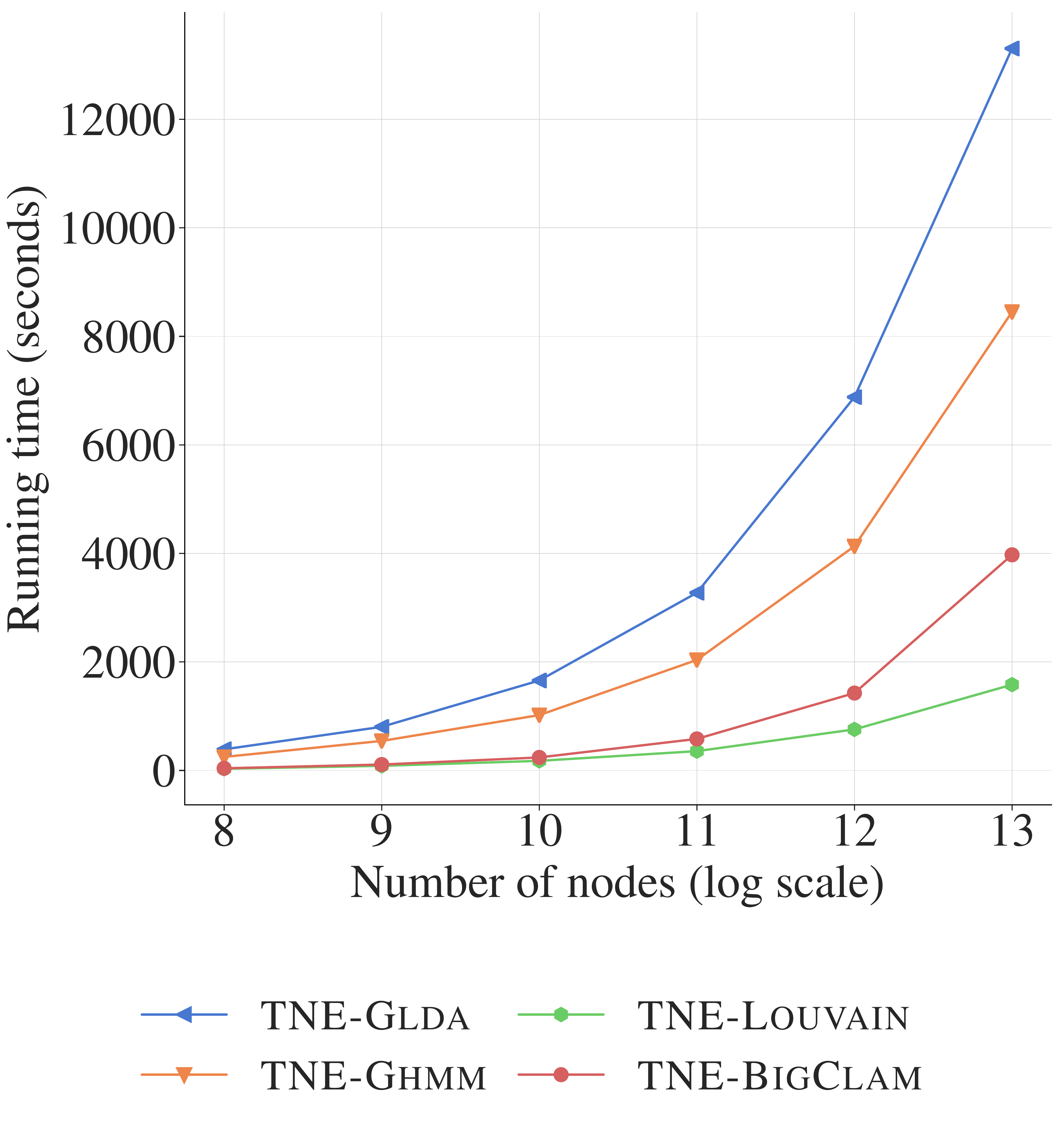}
\caption{Running time of \textsc{TNE} on Erd\"os-R\'enyi graphs of different size.}
\label{fig:running_time}
\end{figure}

In the learning process of topic assignments of nodes with the \textsc{TNE-Glda} model,  we perform collapsed Gibbs sampling \citep{collapsed_gibbs_sampling} for parameter estimation and for inference. MCMC approach is used for the parameter estimation of \textsc{TNE-Ghmm}, with beam sampling for latent sequence resampling steps \citep{beam_sampling}. The number of topics for \textsc{TNE-Glda} is set to $100$ for all networks except \textsl{Citeseer} and \textsl{Cora}, which is set to $150$ and $125$, respectively. The initial number of states for \textsc{TNE-Ghmm} is set to $20$. For the \textsc{M-NMF} algorithm, we have performed parameter tuning for the number of communities on the set of $\{5, 15, 20, 25, 50, 75, 100\}$.

We have performed all the experiments on an Intel Xeon $2.4$GHz CPU server ($32$ Cores) with $60$GB of memory. In order to examine the exact running time of  the \textsc{TNE} variants, we have performed an  experiment on artificially generated Erd\"os-R\'enyi random graphs of varying sizes, ranging from $2^8$ to $2^{13}$ nodes. The running times are reported in Figure \ref{fig:running_time}. As expected, the \textsc{TNE-Louvain} is the most scalable instance, due to the efficient way to infer the community structure.

\begin{table}[H]
\centering
\caption{Node classification for varying training sizes over \textsl{CiteSeer}. For each method, the rows indicates the Micro-$F_1$ and Macro-$F_1$ scores, respectively. The bold value indicates indicate the highest scores.}
\label{tab:classification_citeseer}
\resizebox{0.99\linewidth}{!}{%
\begin{tabular}{rccccccccc}\toprule
\textbf{} & \textbf{10\%} & \textbf{20\%} & \textbf{30\%} & \textbf{40\%} & \textbf{50\%} & \textbf{60\%} & \textbf{70\%} & \textbf{80\%} & \textbf{90\%} \\\midrule
 & 0.517 & 0.548 & 0.566 & 0.576 & 0.584 & 0.590 & 0.595 & 0.591 & 0.592 \\
\multirow{-2}{*}{\textsc{DeepWalk}} & \cellcolor[HTML]{EFEFEF}0.474 & \cellcolor[HTML]{EFEFEF}0.505 & \cellcolor[HTML]{EFEFEF}0.521 & \cellcolor[HTML]{EFEFEF}0.529 & \cellcolor[HTML]{EFEFEF}0.537 & \cellcolor[HTML]{EFEFEF}0.542 & \cellcolor[HTML]{EFEFEF}0.547 & \cellcolor[HTML]{EFEFEF}0.541 & \cellcolor[HTML]{EFEFEF}0.543 \\
 & 0.541 & 0.570 & 0.585 & 0.590 & 0.597 & 0.598 & 0.601 & 0.596 & 0.599 \\
\multirow{-2}{*}{\textsc{Node2Vec}} & \cellcolor[HTML]{EFEFEF}0.494 & \cellcolor[HTML]{EFEFEF}0.525 & \cellcolor[HTML]{EFEFEF}0.537 & \cellcolor[HTML]{EFEFEF}0.542 & \cellcolor[HTML]{EFEFEF}0.549 & \cellcolor[HTML]{EFEFEF}0.550 & \cellcolor[HTML]{EFEFEF}0.553 & \cellcolor[HTML]{EFEFEF}0.548 & \cellcolor[HTML]{EFEFEF}0.551 \\
 & 0.466 & 0.513 & 0.532 & 0.543 & 0.551 & 0.556 & 0.560 & 0.568 & 0.564 \\
\multirow{-2}{*}{\textsc{LINE}} & \cellcolor[HTML]{EFEFEF}0.414 & \cellcolor[HTML]{EFEFEF}0.459 & \cellcolor[HTML]{EFEFEF}0.480 & \cellcolor[HTML]{EFEFEF}0.492 & \cellcolor[HTML]{EFEFEF}0.498 & \cellcolor[HTML]{EFEFEF}0.505 & \cellcolor[HTML]{EFEFEF}0.505 & \cellcolor[HTML]{EFEFEF}0.514 & \cellcolor[HTML]{EFEFEF}0.513 \\
 & 0.219 & 0.228 & 0.256 & 0.267 & 0.277 & 0.293 & 0.299 & 0.300 & 0.320 \\
\multirow{-2}{*}{\textsc{HOPE}} & \cellcolor[HTML]{EFEFEF}0.078 & \cellcolor[HTML]{EFEFEF}0.094 & \cellcolor[HTML]{EFEFEF}0.127 & \cellcolor[HTML]{EFEFEF}0.136 & \cellcolor[HTML]{EFEFEF}0.150 & \cellcolor[HTML]{EFEFEF}0.168 & \cellcolor[HTML]{EFEFEF}0.178 & \cellcolor[HTML]{EFEFEF}0.183 & \cellcolor[HTML]{EFEFEF}0.205 \\
 & \textbf{0.552} & 0.578 & 0.590 & 0.596 & 0.603 & 0.605 & 0.604 & 0.611 & 0.608 \\
\multirow{-2}{*}{\textsc{NetMF}} & \cellcolor[HTML]{EFEFEF}{\color[HTML]{000000} 0.503} & \cellcolor[HTML]{EFEFEF}{\color[HTML]{000000} 0.529} & \cellcolor[HTML]{EFEFEF}{\color[HTML]{000000} 0.542} & \cellcolor[HTML]{EFEFEF}{\color[HTML]{000000} 0.546} & \cellcolor[HTML]{EFEFEF}{\color[HTML]{000000} 0.553} & \cellcolor[HTML]{EFEFEF}{\color[HTML]{000000} 0.554} & \cellcolor[HTML]{EFEFEF}{\color[HTML]{000000} 0.552} & \cellcolor[HTML]{EFEFEF}{\color[HTML]{000000} 0.560} & \cellcolor[HTML]{EFEFEF}{\color[HTML]{000000} 0.554} \\
 & 0.493 & 0.520 & 0.536 & 0.544 & 0.549 & 0.554 & 0.557 & 0.560 & 0.556 \\

\multirow{-2}{*}{\textsc{GEMSEC}} & \cellcolor[HTML]{EFEFEF}{\color[HTML]{000000}  0.452} & \cellcolor[HTML]{EFEFEF}{\color[HTML]{000000}  0.475} & \cellcolor[HTML]{EFEFEF}{\color[HTML]{000000}  0.488} & \cellcolor[HTML]{EFEFEF}{\color[HTML]{000000}  0.493} & \cellcolor[HTML]{EFEFEF}{\color[HTML]{000000}  0.495} & \cellcolor[HTML]{EFEFEF}{\color[HTML]{000000}  0.498} & \cellcolor[HTML]{EFEFEF}{\color[HTML]{000000}  0.502} & \cellcolor[HTML]{EFEFEF}{\color[HTML]{000000}  0.502} & \cellcolor[HTML]{EFEFEF}{\color[HTML]{000000}  0.496} \\
 & 0.386 & 0.428 & 0.443 & 0.458 & 0.458 & 0.466 & 0.467 & 0.472 & 0.471  \\
\multirow{-2}{*}{\textsc{M-NMF}} & \cellcolor[HTML]{EFEFEF}{\color[HTML]{000000} 0.315} & \cellcolor[HTML]{EFEFEF}{\color[HTML]{000000} 0.361} & \cellcolor[HTML]{EFEFEF}{\color[HTML]{000000} 0.379} & \cellcolor[HTML]{EFEFEF}{\color[HTML]{000000} 0.394} & \cellcolor[HTML]{EFEFEF}{\color[HTML]{000000} 0.396} & \cellcolor[HTML]{EFEFEF}{\color[HTML]{000000} 0.404} & \cellcolor[HTML]{EFEFEF}{\color[HTML]{000000} 0.406} & \cellcolor[HTML]{EFEFEF}{\color[HTML]{000000} 0.413} & \cellcolor[HTML]{EFEFEF}{\color[HTML]{000000} 0.407}  \\\midrule
 & 0.535 & 0.578 & 0.593 & 0.605 & 0.610 & 0.616 & 0.616 & 0.621 & 0.618 \\
\multirow{-2}{*}{\textsc{TNE-Glda}} & \cellcolor[HTML]{EFEFEF}0.494 & \cellcolor[HTML]{EFEFEF}0.536 & \cellcolor[HTML]{EFEFEF}0.548 & \cellcolor[HTML]{EFEFEF}0.558 & \cellcolor[HTML]{EFEFEF}0.562 & \cellcolor[HTML]{EFEFEF}0.567 & \cellcolor[HTML]{EFEFEF}0.568 & \cellcolor[HTML]{EFEFEF}0.572 & \cellcolor[HTML]{EFEFEF}0.567 \\
 & 0.526 & 0.557 & 0.568 & 0.575 & 0.583 & 0.584 & 0.583 & 0.592 & 0.594 \\
\multirow{-2}{*}{\textsc{TNE-Ghmm}} & \cellcolor[HTML]{EFEFEF}0.481 & \cellcolor[HTML]{EFEFEF}0.512 & \cellcolor[HTML]{EFEFEF}0.521 & \cellcolor[HTML]{EFEFEF}0.527 & \cellcolor[HTML]{EFEFEF}0.534 & \cellcolor[HTML]{EFEFEF}0.535 & \cellcolor[HTML]{EFEFEF}0.533 & \cellcolor[HTML]{EFEFEF}0.542 & \cellcolor[HTML]{EFEFEF}0.543 \\
 & 0.551 & \textbf{0.589} & \textbf{0.604} & \textbf{0.614} & \textbf{0.619} & \textbf{0.624} & \textbf{0.627} & \textbf{0.632} & \textbf{0.638} \\
\multirow{-2}{*}{\textsc{TNE-Louvain}} & \cellcolor[HTML]{EFEFEF}\textbf{0.506} & \cellcolor[HTML]{EFEFEF}\textbf{0.547} & \cellcolor[HTML]{EFEFEF}\textbf{0.560} & \cellcolor[HTML]{EFEFEF}\textbf{0.570} & \cellcolor[HTML]{EFEFEF}\textbf{0.576} & \cellcolor[HTML]{EFEFEF}\textbf{0.580} & \cellcolor[HTML]{EFEFEF}\textbf{0.582} & \cellcolor[HTML]{EFEFEF}\textbf{0.588} & \cellcolor[HTML]{EFEFEF}\textbf{0.594} \\
 & 0.546 & 0.580 & 0.597 & 0.607 & 0.612 & 0.613 & 0.619 & 0.621 & 0.624 \\
\multirow{-2}{*}{\textsc{TNE-BigClam}} & \cellcolor[HTML]{EFEFEF}0.502 & \cellcolor[HTML]{EFEFEF}0.536 & \cellcolor[HTML]{EFEFEF}0.549 & \cellcolor[HTML]{EFEFEF}0.557 & \cellcolor[HTML]{EFEFEF}0.562 & \cellcolor[HTML]{EFEFEF}0.562 & \cellcolor[HTML]{EFEFEF}0.566 & \cellcolor[HTML]{EFEFEF}0.569 & \cellcolor[HTML]{EFEFEF}0.568\\\bottomrule
\end{tabular}%
}
\end{table}
\begin{table}[H]
\centering
\caption{Node classification for varying training sizes over \textsl{Cora}. For each method, the rows indicates the Micro-$F_1$ and Macro-$F_1$ scores, respectively. The bold value indicates indicate the highest scores.}
\label{tab:classification_cora}
\resizebox{0.99\linewidth}{!}{%
\begin{tabular}{rccccccccc}\toprule
\textbf{} & \textbf{10\%} & \textbf{20\%} & \textbf{30\%} & \textbf{40\%} & \textbf{50\%} & \textbf{60\%} & \textbf{70\%} & \textbf{80\%} & \textbf{90\%} \\\midrule
 & 0.747 & 0.784 & 0.802 & 0.810 & 0.819 & 0.822 & 0.826 & 0.826 & 0.833 \\
\multirow{-2}{*}{\textsc{DeepWalk}} & \cellcolor[HTML]{EFEFEF}0.734 & \cellcolor[HTML]{EFEFEF}0.774 & \cellcolor[HTML]{EFEFEF}0.792 & \cellcolor[HTML]{EFEFEF}0.800 & \cellcolor[HTML]{EFEFEF}0.809 & \cellcolor[HTML]{EFEFEF}0.813 & \cellcolor[HTML]{EFEFEF}0.816 & \cellcolor[HTML]{EFEFEF}0.815 & \cellcolor[HTML]{EFEFEF}0.825 \\
 & 0.769 & 0.799 & 0.815 & 0.824 & 0.831 & 0.835 & 0.839 & 0.842 & 0.841 \\
\multirow{-2}{*}{\textsc{Node2Vec}} & \cellcolor[HTML]{EFEFEF}0.755 & \cellcolor[HTML]{EFEFEF}0.786 & \cellcolor[HTML]{EFEFEF}0.803 & \cellcolor[HTML]{EFEFEF}0.812 & \cellcolor[HTML]{EFEFEF}0.819 & \cellcolor[HTML]{EFEFEF}0.823 & \cellcolor[HTML]{EFEFEF}0.826 & \cellcolor[HTML]{EFEFEF}0.830 & \cellcolor[HTML]{EFEFEF}0.825 \\
 & 0.661 & 0.723 & 0.746 & 0.758 & 0.765 & 0.770 & 0.774 & 0.775 & 0.775 \\
\multirow{-2}{*}{\textsc{LINE}} & \cellcolor[HTML]{EFEFEF}0.642 & \cellcolor[HTML]{EFEFEF}0.713 & \cellcolor[HTML]{EFEFEF}0.736 & \cellcolor[HTML]{EFEFEF}0.747 & \cellcolor[HTML]{EFEFEF}0.755 & \cellcolor[HTML]{EFEFEF}0.759 & \cellcolor[HTML]{EFEFEF}0.762 & \cellcolor[HTML]{EFEFEF}0.766 & \cellcolor[HTML]{EFEFEF}0.764 \\
 & 0.302 & 0.303 & 0.301 & 0.303 & 0.302 & 0.303 & 0.303 & 0.303 & 0.302 \\
\multirow{-2}{*}{\textsc{HOPE}} & \cellcolor[HTML]{EFEFEF}0.066 & \cellcolor[HTML]{EFEFEF}0.067 & \cellcolor[HTML]{EFEFEF}0.067 & \cellcolor[HTML]{EFEFEF}0.067 & \cellcolor[HTML]{EFEFEF}0.067 & \cellcolor[HTML]{EFEFEF}0.067 & \cellcolor[HTML]{EFEFEF}0.068 & \cellcolor[HTML]{EFEFEF}0.070 & \cellcolor[HTML]{EFEFEF}0.072 \\
 & \textbf{0.773} & \textbf{0.807} & \textbf{0.821} & 0.828 & 0.834 & 0.839 & 0.841 & 0.839 & 0.844 \\
\multirow{-2}{*}{\textsc{NetMF}} & \cellcolor[HTML]{EFEFEF}{\color[HTML]{000000} \textbf{0.760}} & \cellcolor[HTML]{EFEFEF}{\color[HTML]{000000} \textbf{0.797}} & \cellcolor[HTML]{EFEFEF}{\color[HTML]{000000} \textbf{0.811}} & \cellcolor[HTML]{EFEFEF}{\color[HTML]{000000} 0.819} & \cellcolor[HTML]{EFEFEF}{\color[HTML]{000000} 0.824} & \cellcolor[HTML]{EFEFEF}{\color[HTML]{000000} 0.830} & \cellcolor[HTML]{EFEFEF}{\color[HTML]{000000} 0.832} & \cellcolor[HTML]{EFEFEF}{\color[HTML]{000000} 0.831} & \cellcolor[HTML]{EFEFEF}{\color[HTML]{000000} 0.835} \\
& 0.599 & 0.643 & 0.670 & 0.691 & 0.707 & 0.720 & 0.728 & 0.740 & 0.738 \\
\multirow{-2}{*}{\textsc{GEMSEC}} & \cellcolor[HTML]{EFEFEF}{\color[HTML]{000000} 0.563} & \cellcolor[HTML]{EFEFEF}{\color[HTML]{000000} 0.616} & \cellcolor[HTML]{EFEFEF}{\color[HTML]{000000} 0.647} & \cellcolor[HTML]{EFEFEF}{\color[HTML]{000000} 0.670} & \cellcolor[HTML]{EFEFEF}{\color[HTML]{000000} 0.688} & \cellcolor[HTML]{EFEFEF}{\color[HTML]{000000} 0.700} & \cellcolor[HTML]{EFEFEF}{\color[HTML]{000000} 0.709} & \cellcolor[HTML]{EFEFEF}{\color[HTML]{000000} 0.722} & \cellcolor[HTML]{EFEFEF}{\color[HTML]{000000} 0.718} \\
 & 0.503 & 0.573 & 0.602 & 0.622 & 0.636 & 0.648 & 0.655 & 0.648 & 0.661 \\
\multirow{-2}{*}{\textsc{M-NMF}} & \cellcolor[HTML]{EFEFEF}{\color[HTML]{000000} 0.433} & \cellcolor[HTML]{EFEFEF}{\color[HTML]{000000} 0.535} & \cellcolor[HTML]{EFEFEF}{\color[HTML]{000000} 0.573} & \cellcolor[HTML]{EFEFEF}{\color[HTML]{000000} 0.596} & \cellcolor[HTML]{EFEFEF}{\color[HTML]{000000} 0.613} & \cellcolor[HTML]{EFEFEF}{\color[HTML]{000000} 0.626} & \cellcolor[HTML]{EFEFEF}{\color[HTML]{000000} 0.634} & \cellcolor[HTML]{EFEFEF}{\color[HTML]{000000} 0.627} & \cellcolor[HTML]{EFEFEF}{\color[HTML]{000000} 0.637} \\\midrule
 & 0.751 & 0.794 & 0.817 & \textbf{0.831} & \textbf{0.840} & \textbf{0.846} & \textbf{0.850} & 0.851 & \textbf{0.856} \\
\multirow{-2}{*}{\textsc{TNE-Glda}} & \cellcolor[HTML]{EFEFEF}0.738 & \cellcolor[HTML]{EFEFEF}0.782 & \cellcolor[HTML]{EFEFEF}0.807 & \cellcolor[HTML]{EFEFEF}\textbf{0.821} & \cellcolor[HTML]{EFEFEF}\textbf{0.829} & \cellcolor[HTML]{EFEFEF}\textbf{0.836} & \cellcolor[HTML]{EFEFEF}\textbf{0.837} & \cellcolor[HTML]{EFEFEF}0.838 & \cellcolor[HTML]{EFEFEF}\textbf{0.841} \\
 & 0.762 & 0.791 & 0.809 & 0.816 & 0.824 & 0.827 & 0.828 & 0.832 & 0.834 \\
\multirow{-2}{*}{\textsc{TNE-Ghmm}} & \cellcolor[HTML]{EFEFEF}0.746 & \cellcolor[HTML]{EFEFEF}0.780 & \cellcolor[HTML]{EFEFEF}0.798 & \cellcolor[HTML]{EFEFEF}0.805 & \cellcolor[HTML]{EFEFEF}0.812 & \cellcolor[HTML]{EFEFEF}0.815 & \cellcolor[HTML]{EFEFEF}0.816 & \cellcolor[HTML]{EFEFEF}0.819 & \cellcolor[HTML]{EFEFEF}0.823 \\
 & 0.765 & 0.797 & 0.815 & 0.826 & 0.835 & 0.843 & 0.846 & \textbf{0.853} & 0.850 \\
\multirow{-2}{*}{\textsc{TNE-Louvain}} & \cellcolor[HTML]{EFEFEF}0.749 & \cellcolor[HTML]{EFEFEF}0.782 & \cellcolor[HTML]{EFEFEF}0.802 & \cellcolor[HTML]{EFEFEF}0.813 & \cellcolor[HTML]{EFEFEF}0.822 & \cellcolor[HTML]{EFEFEF}0.830 & \cellcolor[HTML]{EFEFEF}0.832 & \cellcolor[HTML]{EFEFEF}\textbf{0.839} & \cellcolor[HTML]{EFEFEF}0.836 \\
 & 0.754 & 0.792 & 0.809 & 0.820 & 0.828 & 0.836 & 0.837 & 0.847 & 0.851 \\
\multirow{-2}{*}{\textsc{TNE-BigClam}} & \cellcolor[HTML]{EFEFEF}0.739 & \cellcolor[HTML]{EFEFEF}0.780 & \cellcolor[HTML]{EFEFEF}0.798 & \cellcolor[HTML]{EFEFEF}0.810 & \cellcolor[HTML]{EFEFEF}0.818 & \cellcolor[HTML]{EFEFEF}0.826 & \cellcolor[HTML]{EFEFEF}0.828 & \cellcolor[HTML]{EFEFEF}0.838 & \cellcolor[HTML]{EFEFEF}0.838\\\bottomrule
\end{tabular}%
}
\end{table}
\begin{table}[H]
\centering
\caption{Node classification for varying training sizes over \textsl{DBLP}. For each method, the rows indicates the Micro-$F_1$ and Macro-$F_1$ scores, respectively. The bold value indicates indicate the highest scores.}
\label{tab:classification_dblp}
\resizebox{0.99\linewidth}{!}{%
\begin{tabular}{rccccccccc}\toprule
\textbf{} & \textbf{10\%} & \textbf{20\%} & \textbf{30\%} & \textbf{40\%} & \textbf{50\%} & \textbf{60\%} & \textbf{70\%} & \textbf{80\%} & \textbf{90\%} \\\midrule
 & 0.613 & 0.622 & 0.626 & 0.627 & 0.628 & 0.627 & 0.628 & 0.629 & 0.633 \\
\multirow{-2}{*}{\textsc{DeepWalk}} & \cellcolor[HTML]{EFEFEF}0.545 & \cellcolor[HTML]{EFEFEF}0.552 & \cellcolor[HTML]{EFEFEF}0.555 & \cellcolor[HTML]{EFEFEF}0.556 & \cellcolor[HTML]{EFEFEF}0.556 & \cellcolor[HTML]{EFEFEF}0.555 & \cellcolor[HTML]{EFEFEF}0.556 & \cellcolor[HTML]{EFEFEF}0.557 & \cellcolor[HTML]{EFEFEF}0.559 \\
 & 0.622 & 0.632 & 0.636 & 0.638 & 0.638 & 0.640 & 0.639 & 0.640 & 0.639 \\
\multirow{-2}{*}{\textsc{Node2Vec}} & \cellcolor[HTML]{EFEFEF}0.560 & \cellcolor[HTML]{EFEFEF}0.569 & \cellcolor[HTML]{EFEFEF}0.571 & \cellcolor[HTML]{EFEFEF}0.572 & \cellcolor[HTML]{EFEFEF}0.572 & \cellcolor[HTML]{EFEFEF}0.574 & \cellcolor[HTML]{EFEFEF}0.573 & \cellcolor[HTML]{EFEFEF}0.574 & \cellcolor[HTML]{EFEFEF}0.573 \\
 & 0.603 & 0.613 & 0.618 & 0.619 & 0.621 & 0.621 & 0.623 & 0.623 & 0.623 \\
\multirow{-2}{*}{\textsc{LINE}} & \cellcolor[HTML]{EFEFEF}0.535 & \cellcolor[HTML]{EFEFEF}0.548 & \cellcolor[HTML]{EFEFEF}0.552 & \cellcolor[HTML]{EFEFEF}0.554 & \cellcolor[HTML]{EFEFEF}0.556 & \cellcolor[HTML]{EFEFEF}0.555 & \cellcolor[HTML]{EFEFEF}0.557 & \cellcolor[HTML]{EFEFEF}0.558 & \cellcolor[HTML]{EFEFEF}0.559 \\
 & 0.379 & 0.378 & 0.379 & 0.379 & 0.379 & 0.379 & 0.378 & 0.379 & 0.380 \\
\multirow{-2}{*}{\textsc{HOPE}} & \cellcolor[HTML]{EFEFEF}0.137 & \cellcolor[HTML]{EFEFEF}0.137 & \cellcolor[HTML]{EFEFEF}0.137 & \cellcolor[HTML]{EFEFEF}0.137 & \cellcolor[HTML]{EFEFEF}0.137 & \cellcolor[HTML]{EFEFEF}0.137 & \cellcolor[HTML]{EFEFEF}0.137 & \cellcolor[HTML]{EFEFEF}0.138 & \cellcolor[HTML]{EFEFEF}0.138 \\
 & 0.605 & 0.613 & 0.617 & 0.619 & 0.620 & 0.620 & 0.623 & 0.623 & 0.623 \\
\multirow{-2}{*}{\textsc{NetMF}} & \cellcolor[HTML]{EFEFEF}{\color[HTML]{000000} 0.522} & \cellcolor[HTML]{EFEFEF}{\color[HTML]{000000} 0.528} & \cellcolor[HTML]{EFEFEF}{\color[HTML]{000000} 0.530} & \cellcolor[HTML]{EFEFEF}{\color[HTML]{000000} 0.532} & \cellcolor[HTML]{EFEFEF}{\color[HTML]{000000} 0.531} & \cellcolor[HTML]{EFEFEF}{\color[HTML]{000000} 0.531} & \cellcolor[HTML]{EFEFEF}{\color[HTML]{000000} 0.533} & \cellcolor[HTML]{EFEFEF}{\color[HTML]{000000} 0.533} & \cellcolor[HTML]{EFEFEF}{\color[HTML]{000000} 0.533} \\
 & 0.602 & 0.614 & 0.618 & 0.620 & 0.621 & 0.622 & 0.622 & 0.623 & 0.622 \\
\multirow{-2}{*}{\textsc{GEMSEC}} & \cellcolor[HTML]{EFEFEF}{\color[HTML]{000000} 0.531} & \cellcolor[HTML]{EFEFEF}{\color[HTML]{000000} 0.539} & \cellcolor[HTML]{EFEFEF}{\color[HTML]{000000} 0.541} & \cellcolor[HTML]{EFEFEF}{\color[HTML]{000000} 0.542} & \cellcolor[HTML]{EFEFEF}{\color[HTML]{000000} 0.544} & \cellcolor[HTML]{EFEFEF}{\color[HTML]{000000} 0.544} & \cellcolor[HTML]{EFEFEF}{\color[HTML]{000000} 0.544} & \cellcolor[HTML]{EFEFEF}{\color[HTML]{000000} 0.545} & \cellcolor[HTML]{EFEFEF}{\color[HTML]{000000} 0.543} \\
 & 0.541 & 0.554 & 0.560 & 0.563 & 0.566 & 0.568 & 0.569 & 0.571 & 0.572 \\
\multirow{-2}{*}{\textsc{M-NMF}} & \cellcolor[HTML]{EFEFEF}{\color[HTML]{000000} 0.403} & \cellcolor[HTML]{EFEFEF}{\color[HTML]{000000} 0.423} & \cellcolor[HTML]{EFEFEF}{\color[HTML]{000000} 0.431} & \cellcolor[HTML]{EFEFEF}{\color[HTML]{000000} 0.436} & \cellcolor[HTML]{EFEFEF}{\color[HTML]{000000} 0.439} & \cellcolor[HTML]{EFEFEF}{\color[HTML]{000000} 0.442} & \cellcolor[HTML]{EFEFEF}{\color[HTML]{000000} 0.443} & \cellcolor[HTML]{EFEFEF}{\color[HTML]{000000} 0.444} & \cellcolor[HTML]{EFEFEF}{\color[HTML]{000000} 0.447} \\\midrule
 & 0.617 & 0.627 & 0.631 & 0.633 & 0.634 & 0.635 & 0.635 & 0.636 & 0.638 \\
\multirow{-2}{*}{\textsc{TNE-Glda}} & \cellcolor[HTML]{EFEFEF}0.554 & \cellcolor[HTML]{EFEFEF}0.560 & \cellcolor[HTML]{EFEFEF}0.563 & \cellcolor[HTML]{EFEFEF}0.565 & \cellcolor[HTML]{EFEFEF}0.565 & \cellcolor[HTML]{EFEFEF}0.566 & \cellcolor[HTML]{EFEFEF}0.566 & \cellcolor[HTML]{EFEFEF}0.567 & \cellcolor[HTML]{EFEFEF}0.569 \\
 & 0.619 & 0.629 & 0.632 & 0.634 & 0.635 & 0.636 & 0.636 & 0.636 & 0.634 \\
\multirow{-2}{*}{\textsc{TNE-Ghmm}} & \cellcolor[HTML]{EFEFEF}0.552 & \cellcolor[HTML]{EFEFEF}0.561 & \cellcolor[HTML]{EFEFEF}0.563 & \cellcolor[HTML]{EFEFEF}0.565 & \cellcolor[HTML]{EFEFEF}0.565 & \cellcolor[HTML]{EFEFEF}0.566 & \cellcolor[HTML]{EFEFEF}0.567 & \cellcolor[HTML]{EFEFEF}0.567 & \cellcolor[HTML]{EFEFEF}0.564 \\
 & \textbf{0.625} & \textbf{0.636} & \textbf{0.638} & \textbf{0.641} & 0.642 & 0.641 & 0.642 & 0.643 & \textbf{0.642} \\
\multirow{-2}{*}{\textsc{TNE-Louvain}} & \cellcolor[HTML]{EFEFEF}0.564 & \cellcolor[HTML]{EFEFEF}0.572 & \cellcolor[HTML]{EFEFEF}0.574 & \cellcolor[HTML]{EFEFEF}0.576 & \cellcolor[HTML]{EFEFEF}0.577 & \cellcolor[HTML]{EFEFEF}0.577 & \cellcolor[HTML]{EFEFEF}0.578 & \cellcolor[HTML]{EFEFEF}0.579 & \cellcolor[HTML]{EFEFEF}0.576 \\
 & \textbf{0.625} & 0.635 & \textbf{0.638} & \textbf{0.641} & 0.641 & \textbf{0.643} & \textbf{0.643} & \textbf{0.644} & \textbf{0.642} \\
\multirow{-2}{*}{\textsc{TNE-BigClam}} & \cellcolor[HTML]{EFEFEF}\textbf{0.566} & \cellcolor[HTML]{EFEFEF}\textbf{0.573} & \cellcolor[HTML]{EFEFEF}\textbf{0.576} & \cellcolor[HTML]{EFEFEF}\textbf{0.579} & \cellcolor[HTML]{EFEFEF}\textbf{0.578} & \cellcolor[HTML]{EFEFEF}\textbf{0.580} & \cellcolor[HTML]{EFEFEF}\textbf{0.579} & \cellcolor[HTML]{EFEFEF}\textbf{0.581} & \cellcolor[HTML]{EFEFEF}\textbf{0.579}\\\bottomrule
\end{tabular}%
}
\end{table}

\begin{figure*}
\centering
  \begin{subfigure}[b]{0.3\linewidth}
    \includegraphics[width=\linewidth]{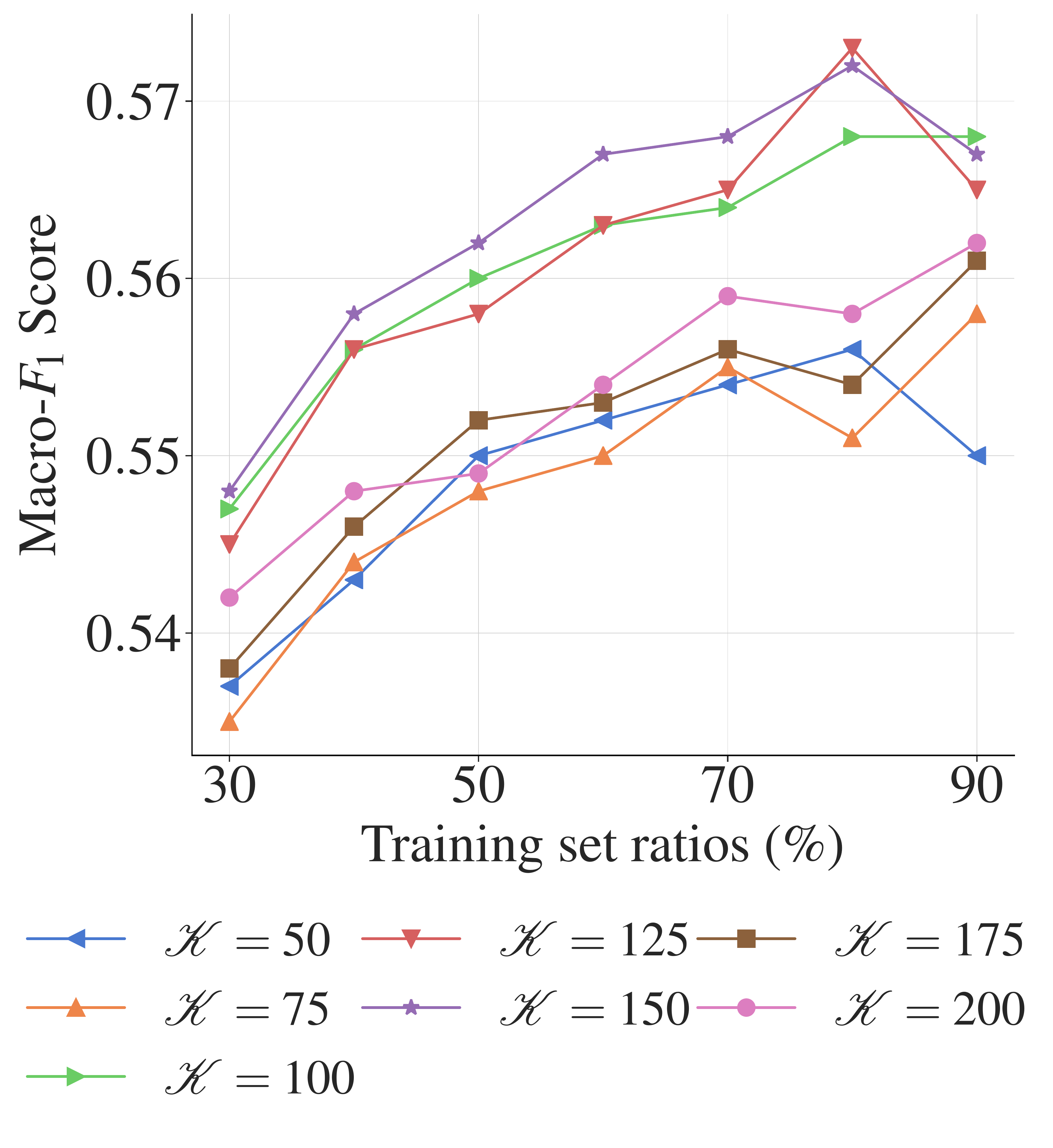}
    \caption{Number of topics on \textsc{TNE-Glda}}
    \label{fig:lda_num_of_comms_analysis}
  \end{subfigure}
  \begin{subfigure}[b]{0.3\linewidth}
     \includegraphics[width=\linewidth]{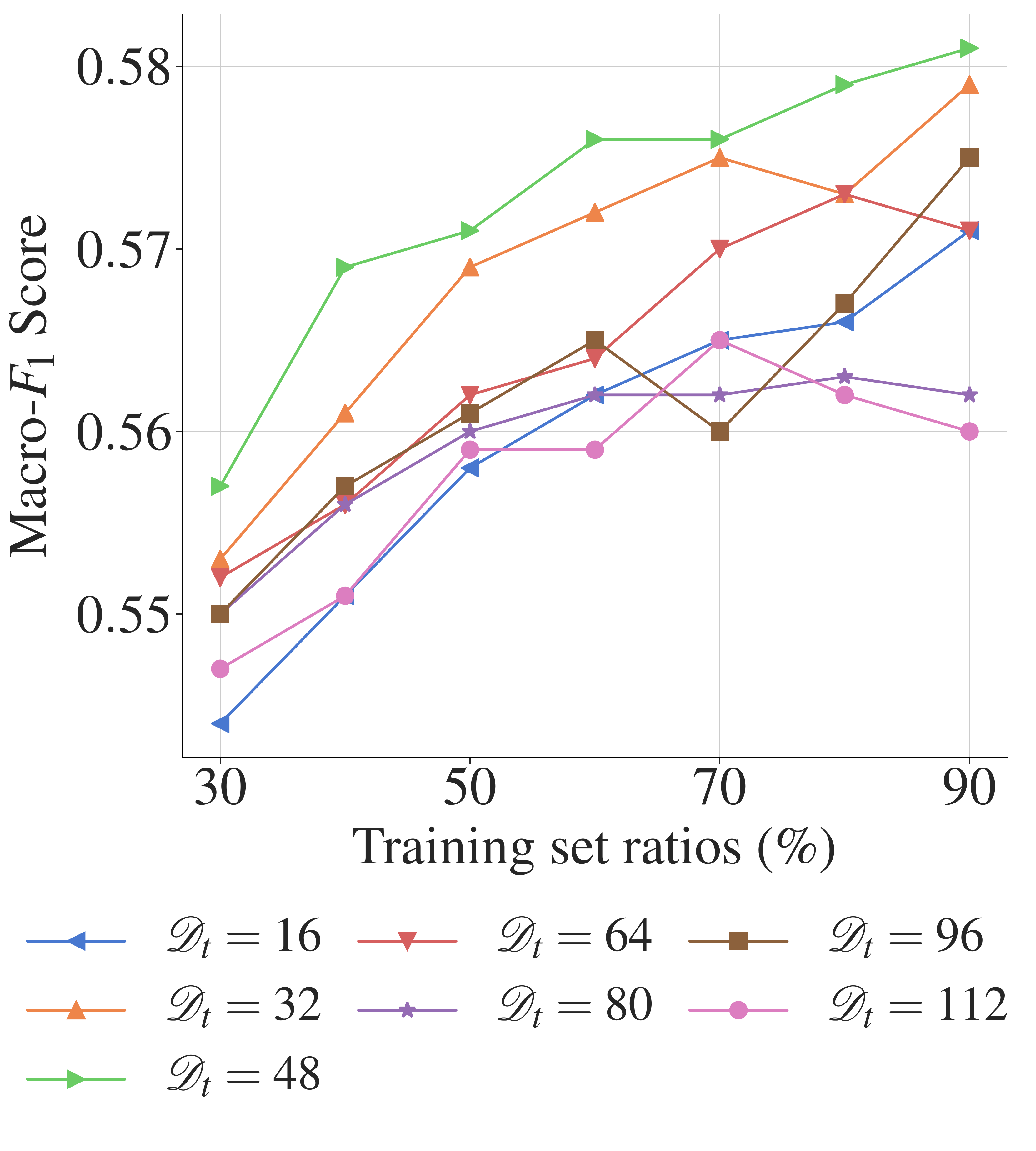}
    \caption{Topic embedding size of \textsc{TNE-Louvain}}
    \label{fig:dim_analysis_lda_louvain}
  \end{subfigure}
  \caption{Parameter sensitivity of \textsc{TNE-Glda} and \textsc{TNE-Louvain} models.}
\end{figure*}

\subsection{Node classification}

\paragraph{Experimental setup} In the node classification task, nodes are associated with labels and the goal is to predict the correct labels, observing only certain fraction of the network.  We split the collection of feature vectors into training and tests sets, and apply an one-vs-rest logistic regression classifier with $L2$ regularization for optimization. In order to provide more reliable experimental results, the same procedure is repeated for $50$ times. 

\paragraph{Results} The detailed results for the different instances of TNE framework as well as for the baseline methods are provided in Tables \ref{tab:classification_citeseer}, \ref{tab:classification_cora} and \ref{tab:classification_dblp}. Experiments are reported for different ratios of training data ($10\%$-$90\%$). For each model, the first row corresponds to Micro-$F_1$ scores, while the second one to Macro-$F_1$ scores.

As we can observe, the \textsc{TNE-Louvain} and \textsc{TNE-BigClam} models perform quite well, outperforming most of the baseline methods for almost all different training ratios over \textsl{CiteSeer} and \textsl{DBLP} datasets. In the case of the \textsl{Cora} network, \textsc{TNE-Glda} performs better especially when the train-test split of the dataset is quite balanced. The overall better performance of the two network structure-based instances \textsc{TNE-Louvain} and \textsc{TNE-BigClam} over random walk-based models (e.g., \textsc{TNE-Glda}), can possibly be explained by the way that latent communities are extracted. While with \textsc{TNE-Glda} and \textsc{TNE-Ghmm} we are able to utilize random walks for both node and topic representations, it seems that those random walk are not expressive enough to recover the clustering structure as effectively as algorithms tailored to this task, such as \textsc{BigClam} and \textsc{Louvain}. With respect to the  baseline methods, we can observe that \textsc{NetMF} performs well especially in the case of the \textsl{Cora} network.

\subsection{The effect of number of topics and topic embedding sizes}
We have further analyzed the effect of the chosen number of topics $\mathcal{K}$ on the performance of the \textsc{TNE-Glda} model on the \textsl{CiteSeer} network.  As shown in Figure \ref{fig:lda_num_of_comms_analysis}, the increase in the number of topics makes positive contribution to the classification performance up to a certain level for \textsc{TNE-Glda} model, reaching its highest $F_1$-score for $\mathcal{K}=150$. The chosen number of topics seems to be a more crucial parameter especially for larger training data sizes.

The size of the topic embedding vector learned by Eq. \eqref{eq: community_emb_obj_func} is another factor affecting the performance of TNE. In this paragraph, we examine the influence of the topic embedding size on the \textsc{TNE-Louvain} model over 
\textsl{Citeseer} network. In the experiment, we have fixed the total embedding length to $\mathcal{D}_n+\mathcal{D}_t$ to $128$, and we vary the length of topic ($\mathcal{D}_t$) and node ($\mathcal{D}_n$) embeddings. As it can be observed in Figure \ref{fig:dim_analysis_lda_louvain}, for small node embedding sizes, the scores diminish drastically since topic embeddings alone  do not convey sufficient information to effectively represent the nodes 
On the contrary, when we accurately balance the lengths of topic and node embeddings (for $\mathcal{D}_t=48$ in this particular case), the contribution of topic-enhanced embeddings over purely node embeddings is evident.

\subsection{Link Prediction}
\paragraph{Experimental setup} In the link prediction task, we have limited access to the edges of the network, and our goal is to predict the missing (unseen) edges between nodes. We divide the edge set  of a given network into two parts to form training and test sets, by randomly removing $50\%$ of  the edges (the network  remains connected during the process). The removed edges are later used as positive samples in the test set. The same number of node pairs which does not exist in the initial network is chosen to obtain negative samples for each training and test sets. The node embedding vectors are converted into edge features by using the \textit{Hadamard} product. We perform experiments using the logistic regression classifier with $L2$ regularization.

\paragraph{Results} Table \ref{tab:link_prediction} shows the area under curve (AUC) scores for link prediction. The different  instances of TNE are able to outperform the baselines except \textsl{Gnutella}, and the \textsc{TNE-Ghmm} model shows the best performance across multiple datasets.

\begin{table}[]
\centering
\caption{Area Under Curve (AUC) scores for the link prediction task.}
\label{tab:link_prediction}
\resizebox{0.99\linewidth}{!}{%
\begin{tabular}{rcccccccc}
\toprule
 & \textsl{Citeseer} & \textsl{Cora} & \textsl{DBLP} & \textsl{AstroPh} & \textsl{HepTh} & \textsl{Facebook} & \textsl{Gnutella} &  \\\midrule
\textsc{DeepWalk} & 0.770 & 0.739 & 0.919 & 0.911 & 0.843 & 0.980 & 0.559 &  \\
\textsc{Node2Vec} & 0.780 & 0.757 & 0.954 & 0.969 & 0.896 & 0.992 & 0.605 &  \\
\textsc{LINE} & 0.717 & 0.686 & 0.933 & 0.971 & 0.854 & 0.986 & 0.569 &  \\
\textsc{HOPE} & 0.744 & 0.712 & 0.873 & 0.931 & 0.836 & 0.975 & 0.636 &  \\
\textsc{NetMF} & 0.763 & 0.794 & 0.912 & 0.885 & 0.869 & 0.985 & 0.621 &  \\
\textsc{GEMSEC} & 0.748 & 0.733 & 0.920 & 0.816 & 0.823 & 0.473 & 0.661 &  \\
\textsc{M-NMF} & 0.745 & 0.726 & 0.894 & 0.947 & 0.864 & 0.981 & \textbf{0.683}         &  \\\midrule
\textsc{TNE-Glda} & \textbf{0.809} & 0.775 & 0.958 & 0.977 & 0.903 & \textbf{0.993} & 0.629 &  \\
\textsc{TNE-Ghmm} & 0.793 & \textbf{0.806} & \textbf{0.959} & \textbf{0.979} & \textbf{0.908} & \textbf{0.993} & 0.663 &  \\
\textsc{TNE-Louvain} & \textbf{0.809} & 0.780 & 0.959 & 0.977 & 0.905 & \textbf{0.993} & 0.637 &  \\
\textsc{TNE-BigClam} & 0.792 & 0.767 & 0.958 & 0.977 & 0.904 & \textbf{0.993} & 0.631 & \\\bottomrule
\end{tabular}%
}
\end{table}

\subsection{Further Empirical Analysis of \textsc{TNE}}

\begin{figure}[t]
\centering
    \includegraphics[width=0.85\columnwidth]{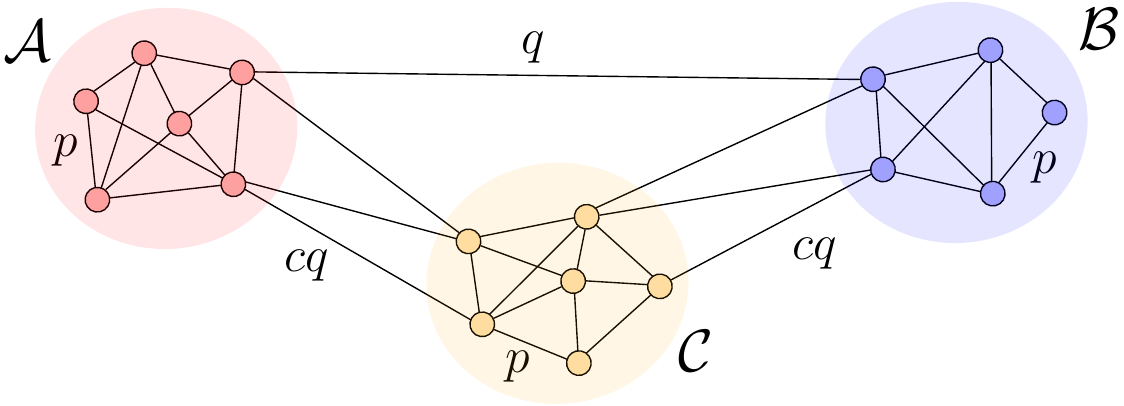}
\caption{Illustration of link sampling in the stochastic block model.}%
\label{fig:artificial_networks}
\end{figure}

Here, we further analyze the behaviour of the various \textsc{TNE} instances on artificially generated networks, focusing on controlled classification experiments. We generate random graphs by following a similar approach as  in \citep{berberidis2018node}. In particular, we use the \textit{stochastic block model} to construct graphs that consist of three clusters $\mathcal{A}$, $\mathcal{B}$ and $\mathcal{C}$, each one of size $1,000$ nodes.  Figure \ref{fig:artificial_networks} provides a schematic representation of the process. An intra-community link is added with probability $p$, while an edge between  communities $\mathcal{A}$ and $\mathcal{B}$ is added with probability $q$. We use an additional parameter $c$ in order to introduce asymmetry for inter-community links for the community pairs $\mathcal{A}-\mathcal{C}$ and $\mathcal{B}-\mathcal{C}$. We have constructed three networks ($\mathcal{G}_1$, $\mathcal{G}_2$, $\mathcal{G}_3$) for parameters $p=[0.06, 0.065, 0.07]$, $q=[0.04, 0.035, 0.003]$ and $c=[1.25, 1.429, 1.667]$, respectively. In our classification experiments, node labels correspond to community assignments. 

Table \ref{tab:artificial_table} provides the Micro-$F_1$ scores for different walk lengths on  $\mathcal{G}_1$, $\mathcal{G}_2$ and $\mathcal{G}_3$, with $40\%$ of the datasets used as training set. We perform uniform random walks following \textsc{DeepWalk}'s strategy, and the instances of \textsc{TNE} are fed with the same node sequences. As we can observe, the performance of the models increases depending on the walk length, since by increasing the number of center-context node pairs improves the ability of the models to capture the structural properties of the graph. Comparing the extreme cases of $\mathcal{G}_1$ and $\mathcal{G}_3$,  the latter shows more distinguishable community structure. Therefore, TNE has better performance on $\mathcal{G}_3$ since the community detection algorithms are able to correctly assign topic-community labels. Among them, \textsc{TNE-Louvain} is the best performing instance. It detects good modularity communities from the graph structure itself, and thus, contrary to \textsc{TNE-Glda} and \textsc{TNE-Ghmm},  it does not rely on the the generated random walk sequences. Furthermore, it is more successful to find the clusters of the stochastic block model graphs compared to \textsc{TNE-BigClam}, since \textsc{BigClam} focuses on overlapping communities that do not appear on these artificially generated graphs.

\begin{table}
\centering
\caption{Micro-$F_1$ classification accuracy on  the stochastic block model,  with $40\%$ training set ratio.}
\label{tab:artificial_table}
\resizebox{0.99\linewidth}{!}{%
\begin{tabular}{rccccccccc}
\toprule
\multicolumn{1}{c}{} & \multicolumn{3}{c}{$\mathbf{\mathcal{G}_1}$} & \multicolumn{3}{c}{$\mathbf{\mathcal{G}_2}$} & \multicolumn{3}{c}{$\mathbf{\mathcal{G}_3}$}\\\midrule
 \textbf{Walk length,} $\mathbf{\mathcal{L}}$ & $10$ & $50$ & $100$ & $10$ & $50$ & $100$ & $10$ & $50$ & $100$ \\\cmidrule(lr){1-1}\cmidrule(lr){2-4}\cmidrule(lr){5-7}\cmidrule(lr){8-10}
\textsc{DeepWalk}  & 0.708 & 0.721 & 0.720 & 0.859 & 0.856 & 0.861 & 0.914 & 0.931 & 0.936 \\
\textsc{TNE-Glda} & 0.708 & 0.750 & 0.759 & 0.860 & 0.875 & 0.876 & 0.913 & 0.938 & 0.944 \\
\textsc{TNE-Ghmm}  &  0.707 & 0.752 & 0.762 & 0.861 & 0.877 & 0.875 & 0.916 & 0.938 & 0.944  \\
\textsc{TNE-Louvain}  & \textbf{0.717} & \textbf{0.769} & \textbf{0.771} & \textbf{0.864} & \textbf{0.882} & \textbf{0.877} & \textbf{0.964} & \textbf{0.964} & \textbf{0.969}  \\
\textsc{TNE-BigClam}  &  0.710 & 0.749 & 0.757 & 0.856 & 0.873 & 0.872 & 0.912 & 0.933 & 0.940 \\\bottomrule
\end{tabular}%
}
\end{table}

\section{Conclusions and Future Work} \label{sec: conclusions}
In this paper, we have proposed TNE, a topic-aware family of models for NRL. TNE takes advantage of the  latent community structure of  graphs, leading to the concept of topical node embeddings. We have examined four  instances of the proposed model, that either use random walks to learn topic representations or rely on well-known community detection algorithms.  TNE is capable of producing enriched   representations, compared to traditional random walk-based approaches or matrix factorization-based models,  leading to improved performance results in the tasks of node classification and link prediction.

\par As future work, we are interested to  extend  TNE towards utilizing the hierarchical community structure of real-world graphs, learning hierarchical node embeddings. Furthermore, we plan to examine the robustness of the learning representations with respect to changes on the structure of the graph.

\paragraph{\normalfont \textbf{Acknowledgments}}
 We thank the anonymous reviewers for the constructive comments.

\bibliographystyle{model2-names}
\bibliography{main}

\end{document}